\documentclass{article}

\usepackage{microtype}
\usepackage{graphicx}
\usepackage{subfigure}
\usepackage{booktabs} 

\usepackage{hyperref}

\usepackage[accepted]{icml2023}

\usepackage{amsmath}
\usepackage{amssymb}
\usepackage{mathtools}
\usepackage{amsthm}

\usepackage[capitalize,noabbrev]{cleveref}

\theoremstyle{plain}
\newtheorem{theorem}{Theorem}[section]

\theoremstyle{definition}

\theoremstyle{remark}

\usepackage[textsize=tiny]{todonotes}

\usepackage{amsmath,amsfonts,bm}

\def\eqref#1{equation~\ref{#1}}

\def\1{\bm{1}}

\DeclareMathAlphabet{\mathsfit}{\encodingdefault}{\sfdefault}{m}{sl}
\SetMathAlphabet{\mathsfit}{bold}{\encodingdefault}{\sfdefault}{bx}{n}

\def\gE{{\mathcal{E}}}

\def\gG{{\mathcal{G}}}

\def\gL{{\mathcal{L}}}

\def\gV{{\mathcal{V}}}

\newcommand{\sigmoid}{\sigma}

\usepackage{hyperref}
\usepackage{url}
\usepackage{graphicx}
\usepackage{enumitem}

\include{cmd}

\icmltitlerunning{\ours: Slot-based Message Passing for Heterogeneous Graphs}

\begin{document}

\twocolumn[
\icmltitle{\ours: Slot-based Message Passing for Heterogeneous Graph Neural Network}



\icmlsetsymbol{equal}{*}

\begin{icmlauthorlist}
\icmlauthor{Ziang Zhou}{polyu}
\icmlauthor{Jieming Shi}{polyu}
\icmlauthor{Renchi Yang}{hkbu}
\icmlauthor{Yuanhang Zou}{tencent}
\icmlauthor{Qing Li}{polyu}
\end{icmlauthorlist}

\icmlaffiliation{polyu}{Department of Computing, The Hong Kong Polytechnique Univesity}
\icmlaffiliation{hkbu}{Department of Computer Science, Hong Kong Baptist University}
\icmlaffiliation{tencent}{Tencent}

\icmlcorrespondingauthor{Jieming Shi}{jieming.shi@polyu.edu.hk}

\icmlkeywords{Machine Learning, ICML}

\vskip 0.3in
]



\printAffiliationsAndNotice{}  

\begin{abstract}
Heterogeneous graphs are ubiquitous to model complex data. 
There are urgent needs on powerful heterogeneous graph neural networks to effectively support important applications. 
We identify a potential semantic mixing issue in existing message passing processes, where the  representations of the neighbors of a node $v$ are forced to be transformed to the feature space of $v$ for aggregation, though the neighbors are in different types. 
That is, the semantics in different node types are entangled together into node $v$'s representation.
To address the issue, we propose \ours with  separate message passing processes in slots, one for each node type, to maintain the representations in their own  node-type feature spaces.
Moreover, in a slot-based message passing layer,  we design an attention mechanism for effective slot-wise message aggregation. 
Further, we develop a slot attention technique after the last layer of \ours, to learn the importance of different slots in downstream tasks.
Our analysis indicates that the slots in \ours can preserve different  semantics in various feature spaces.
The superiority of  \ours is evaluated against 13 baselines on 6 datasets for node classification and link prediction. 
Our code is  at \url{https://github.com/scottjiao/SlotGAT_ICML23/}.
\end{abstract}

\section{Introduction}\label{sec:intro}

\begin{figure}[!t]
     \centering
        \includegraphics[width=0.98\columnwidth]{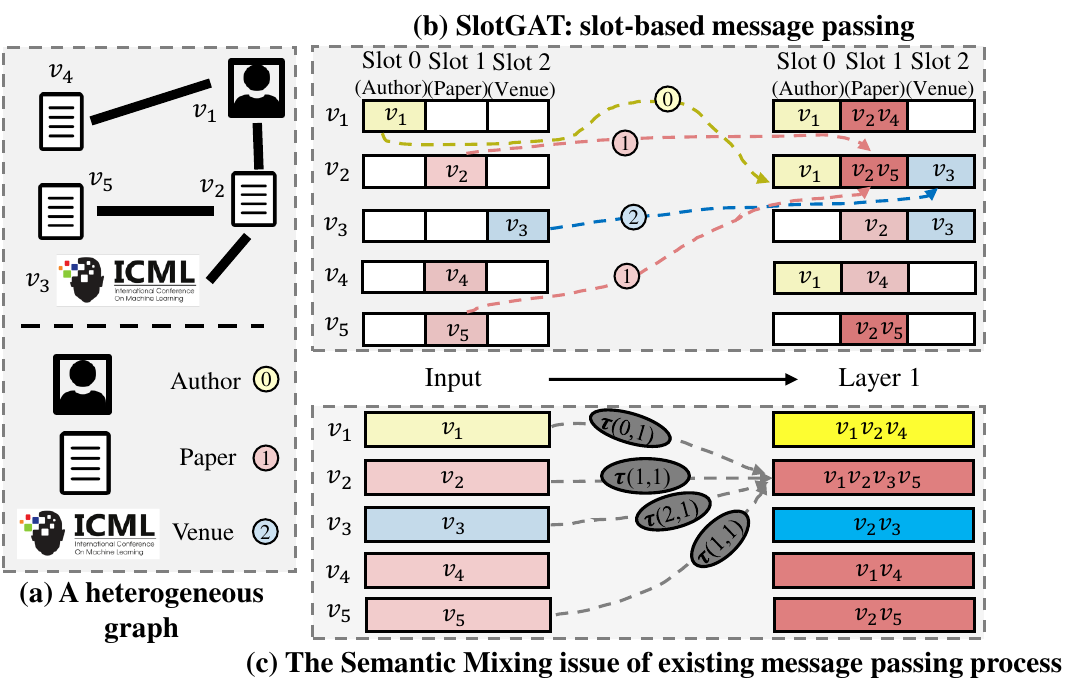}
       \caption{ (a) A heterogeneous graph. (b) The slot-based message passing in the proposed \ours: every node has 3 slots, corresponding to 3 node-type feature spaces. 
       For example, in the input layer, $v_2$'s slot 1 is initialized by its features since $v_2$ is in type 1, while slots 0 and 2 of $v_2$ are empty. The message passing in \ours is slot specific (colored dashed arrows),
       \eg neighboring slot messages passed to the slots of the same node type on node $v_2$ in Layer 1. 
       (c) The semantic mixing issue: every node maintains a single representation; in Layer 1, $v_2$ aggregates message from $v_1$ by firstly applying transformation $\tau(0,1)$ to convert $v_1$'s representation in type 0 to the feature space of $v_2$ in type 1, which mixes the two feature spaces.}
        \label{fig:example}
\end{figure}

Heterogeneous graphs with node and edge types  \cite{Hu:KDD20GPT,dong2017metapath2vec,yang2020multisage}, are ubiquitous in many real applications, \eg protein prediction~\citep{fout2017protein}, recommendation~\citep{fan2019recommendation,yang2020session}, social analysis~\citep{qiu2018social,li2019encoding}, and traffic prediction~\citep{guo2019attention}.
Figure \ref{fig:example}(a) displays  a heterogeneous academic graph with 5 nodes in 3 types, \ie author, paper, venue. Usually edge type is related to the node types of the  two ends of an edge. For instance, the edge between $v_1$ (author) and $v_2$ (paper) is an authorship edge;  $v_2$ and $v_5$ (paper) have a citation edge. 

Heterogeneous graphs have attracted great research attentions.
Conventional graph neural networks (GNNs) \citep{hamilton2017inductive,klicpera2018predict,gcn,velickovic2017graph,monti2017geometric,liu2020towards}  can be applied by regarding heterogeneous graphs as homogeneous ones. GNNs are good enough, but still ignore  the heterogeneous semantics   \citep{lv2021simple}.
Hence, several heterogeneous graph neural networks (HGNNs) are developed \citep{han,gtn,rshn,hetgnn, magnn, hu2020heterogeneous, hong2020attention, schlichtkrull2018modeling, cen2019representation, kgcn, kgnnls, wang2019kgat, lv2021simple, space4HGNN}. 
Some HGNNs rely on meta-paths, either predefined manually \cite{han,magnn} or learned automatically \cite{gtn}, which incur extra costs. 
Then there are studies of relation oriented HGNNs   \cite{rshn,schlichtkrull2018modeling}, HGNNs with random walks  and RNNs \citep{hetgnn}, empowering GNNs for HGNNs \citep{lv2021simple},  etc.
\citet{space4HGNN}   define  a unified design space
for HGNNs. 

It is known  that different node types have different semantics, and naturally should have different impacts~\citep{han,lv2021simple,hu2020heterogeneous}.
However, when aggregating the messages from different types of nodes, the semantics are mixed in existing work, which may  hamper the effectiveness  \cite{han,gtn,rshn,magnn}.
Specifically, in existing message passing layers, when a node $v$ aggregates messages from its neighbors  $u$ that may be in   different node type, they force the representation of $u$ in another type-specific feature space to be transformed to the feature space of $v$'s type, and then aggregate the transformed message to $v$.
We argue that this forced transformation mixes different feature spaces, making representations entangled with each other.
This semantic mixing issue is illustrated in Figure \ref{fig:example}(c). 
Given the heterogeneous graph in the figure,  every node is with a \textit{single} initial representation as input. Then in Layer 1, when node $v_2$ (paper type id 1) aggregates the message from its neighbor $v_1$ (author type id 0)  in a different node type, it first applies a transformation $\tau(0,1)$ to convert the   representation of $v_1$ from author's feature space to paper's feature space, and then aggregates it to paper $v_2$'s representation. Similar forced transformation  also exists when aggregating $v_3$ (venue) to $v_2$ (paper) in Layer 1. 
Instead of converting all   $v$'s neighboring representations to  $v$'s feature space, if we     separate and preserve the impacts of different node-type features to $v$, we could learn more effective representations.

Hence, we propose \ours, which has separate message \textit{slots}   of different node types (\ie feature spaces), with dedicated attention mechanisms to measure the importance of slots. 
\ours conducts \textit{slot-based message passing processes} to alleviate the semantic mixing issue.
We explain the idea of slot-based message passing in Figure \ref{fig:example}(b).
For a node $v$, we maintain a slot for every node type in the heterogeneous graph, \eg every node with 3 slots in Figure \ref{fig:example}(b) for author, paper, venue types, respectively.
When initializing the input slot representations of $v_2$ (paper), the slot for paper type (slot 1) is initialized by $v_2$'s features, while other slots (slots 0, 2) of $v_2$ for author and venue are initialized as empty. 
In Layer 1, when aggregating neighbors to $v_2$, the slot 0 message from $v_1$ is aggregated to the corresponding slot 0 of $v_2$, without mixing with the representations in other slots (\ie other node types).
The slot 2 message from $v_3$ (venue) is aggregated to the respective slot 2 of $v_2$. Obviously, though $v_2$ is in paper type, it can maintain slot representations of other node types, compared with existing methods in  Figure \ref{fig:example}(c) where $v_2$ mixes the representations of $v_1$, $v_2$, $v_3$, $v_5$ in different node types.
In the layers  of \ours, we also design an \textit{attention-based aggregation mechanism} that considers all slot representations of a node and its neighbors as well as edge type representations for aggregation.
In downstream tasks, to effectively integrate the slot representations of a node, after the last layer of \ours,  we develop a \textit{slot attention} technique to learn the importance of different slots. 
Our analysis indicates that \ours can preserve different   semantics into slots.

Compared with numerous existing methods, \ours achieves superior performance on all datasets under various evaluation metrics. We also conduct model analysis, including statistical significance test, visualization, and ablation analysis, to analyze the effectiveness of \ours.

We summarize our contributions as follows: 
\begin{itemize}[topsep=-2pt,leftmargin=*]
  \setlength\itemsep{-1.8pt}
\item We propose a novel slot-based heterogeneous graph neural network model \ours for heterogeneous graphs.
\item We develop a new slot-based message passing mechanism, so that semantics of different node-type feature spaces do not entangle with each other. 
\item We further design attention mechanisms to effectively process the representations in slots.
\item Extensive experiments demonstrate the superiority of \ours on real-world heterogeneous graphs.

\end{itemize}

\section{Preliminary}\label{sec:pre}

A heterogeneous graph is  $\mathcal{G}=\left(\cV,\gE,\phi,\psi\right)$, where $\cV$ is the set of nodes, $\gE$ is the set of edges, $\phi$ is a node type mapping function, and $\psi$ is an edge type mapping function. 
Every node $v$ has node type $\phi(v)$, and every edge $e=(v,u)$ has edge type $\psi(e)$, also denoted as $\psi(v,u)$.
Let $\Phi$ and $\Psi$ be the set of all node types and the set of all edge types respectively in graph $\mathcal{G}$.
Assume that node types in $\Phi$ are represented by consecutive integers starting from $0$, and denote $t$ as the $t$-th node type.
Let $n$ be the number of nodes, \ie $n=|\cV|$, and $m$ be the number of edges, \ie $m=|\gE|$. 
For a heterogeneous graph $|\Psi|+|\Phi|>2$.

A node $v$ has a feature vector $\bx_v$.
For node type $t\in \Phi$, all type-$t$ nodes $\cv\in\{\cv\in\gV|\phi(\cv)=t\}$ have the same feature dimension $d_0^{t}=d_{0}^{\phi(\cv)}$, \ie $\bx_{\cv}\in\real^{d_{0}^{\phi(\cv)}}$. Nodes of different types can have different feature dimensions \cite{lv2021simple}.

\begin{figure*}[!t]
\centering
    \includegraphics[width=0.92\textwidth]{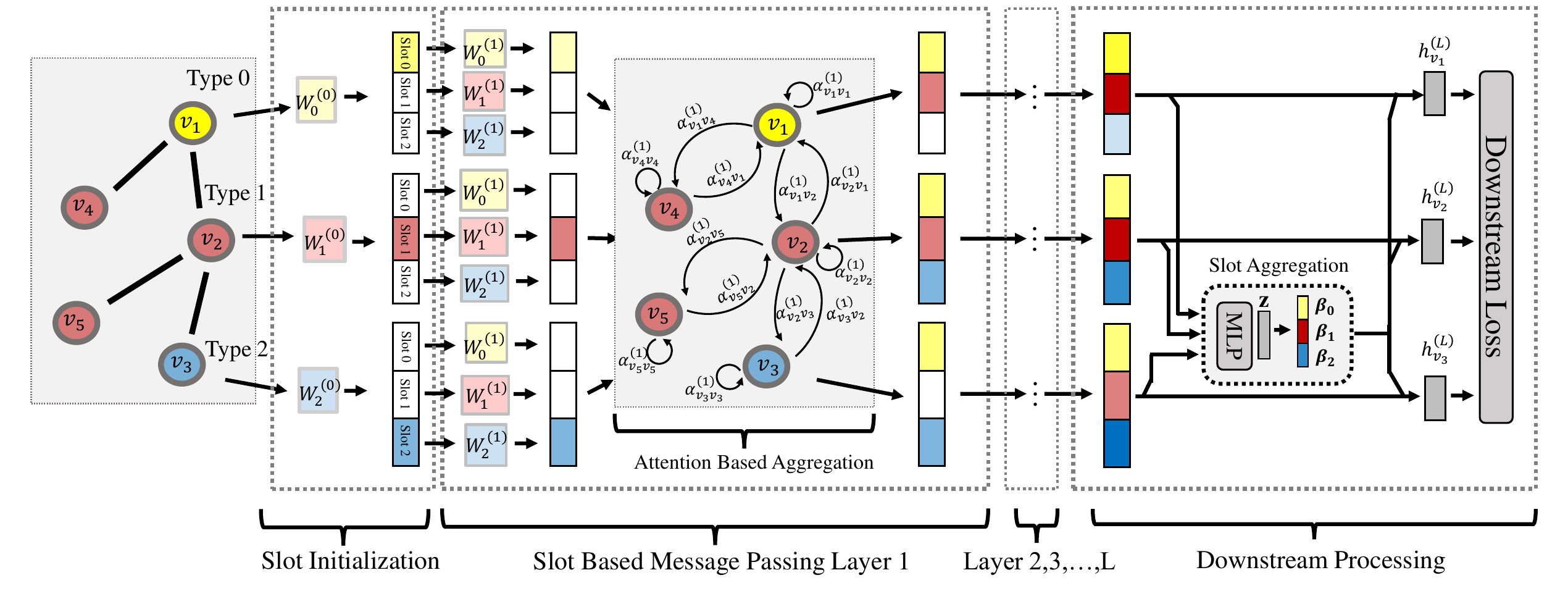}
    \caption{
    The \ours architecture. (i) \ours initializes every node with multiple slots corresponding to all node types (3 types in the example), and the slot for the node's type is initialized by its {transformed} features, while the other slots are empty. (ii) In a slot-based message passing layer, slot representations are transformed and propagated separately, with an attention based aggregation technique.   
    (iii) After the last $L$-th layer, \ours includes a slot attention technique to integrate slots for downstream tasks.  }
\label{fig:arch}
\end{figure*}

\section{Related Work}\label{sec:related}

Homogeneous GNNs handle graphs without node/edge types \citep{gcn,liu2020towards,bruna2013spectral,defferrard2016convolutional,velickovic2017graph,monti2017geometric,chen2020simple,klicpera2018predict}. GCN~\citep{gcn} simplifies spectral networks on graphs~\citep{bruna2013spectral} into its GNN form. GAT \citep{velickovic2017graph} introduces self-attention to GNNs.
Then, there is a plethora of HGNNs   to handle heterogeneous graphs.
Meta-paths are used in
\cite{han,magnn,gtn}.
HAN \cite{han} consists of a hierarchical attention mechanism to capture node-level importance between nodes and semantic-level importance of meta-paths. 
MAGNN~\cite{magnn} is an enhanced method with several meta-path encoders to encode all the information along meta-paths.
Both MAGNN and HAN require meta-paths generated manually.
Graph transformer network (GTN) \cite{gtn} can automatically learn meta-paths  with  graph transformation layers.
For heterogeneous graphs with many edge types, meta-path based methods are not easily applicable, due to the high cost on obtaining meta-paths. 
There are methods treating heterogeneous graphs as relation graphs to develop graph neural network methods. 
RSHN is a relation structure-aware HGNN \cite{rshn}, which builds coarsened line graph to get edge features and adopts a Message Passing Neural Network (MPNN)~\cite{gilmer2017neural} to propagate node and edge features.
RGCN~\cite{schlichtkrull2018modeling} splits a heterogeneous graph to multiple subgraphs by building an independent adjacency matrix for each edge type.  
Furthermore, HGT~\cite{hu2020heterogeneous} is a graph transformer model to handle large heterogeneous graphs with heterogeneous subgraph sampling techniques.
 HetGNN \cite{hetgnn} uses random walks   to sample fixed-size neighbors for nodes in different node type, and then applies RNNs for representation learning.
As shown in experiments, HetGNN is inferior to our method.
HetSANN~\cite{hong2020attention} contains type-specific graph attention layers to aggregate local information.
In experiments,   \ours is better than HetSANN.
Recently, Lv \etal \citep{lv2021simple} develop simpleHGN with several techniques on homogeneous GNNs to handle heterogeneous graphs.
Space4HGNN \cite{space4HGNN} defines a unified design space for HGNNs, to exhaustively evaluate the combinations of many techniques. {\citet{disenHAN} propose a disentangled  mechanism in which  an aspect of a target node  aggregates semantics from all neighbors for recommendation, regardless of node types. Contrarily, we learn separate semantics in different node type feature spaces, which is different.}

\section{The \ours Method}\label{sec:method}

Figure \ref{fig:arch} shows the architecture of \ours, consisting of  node-type slot initialization, slot-based message passing layers with attention based aggregation, and a slot attention module. 

Given a heterogeneous graph $\mathcal{G}$, \ours creates $|\Phi|$ slots for every node  (3 slots per node in Figure \ref{fig:arch}). 
The $t$-th slot of a node $v$ represents the semantic representation of $v$ with respect to node type $t$. 
For initialization, as in Figure \ref{fig:arch}, if slot $t$ of $v$ (\eg $v_3$) corresponds to the node type of $v$, $\phi(v)$ (\eg $t=2$), then  slot $t$ is initialized by node $v$'s features $\mathbf{x}_v$ via a linear transformation. All other slots of $v$ are initialized as zero (\eg slots 0, 1 of $v_3$ in Figure \ref{fig:arch}).

Then within a slot-based message passing layer of \ours, given a node $v$, its neighbors transform and pass slot-wise messages to it. Compared with existing methods that only pass \textit{one} message from a neighbor $u$ to $v$, \ours passes $|\Phi|$ \textit{slot messages separately} to $v$, and these messages do not mix with each other in intermediate layers. As illustrated in Figure \ref{fig:arch}, in  Layer $l=1$,  for any neighbor $u$ of $v$, we apply transformation by $\mathbf{W}_t^{(l)}$ independently on each slot $t$ of $u$, and then pass the transformed slot messages to $v$ for slot-wise aggregation with an attention mechanism that leverages slot representations and edge types to compute attention scores,   elaborated in Section \ref{sec:messagePassing}.
Remark that  the slot $t$ of $v$ preserves the type-$t$ specific semantics delivered from the graph to $v$, regardless whether $v$ is in type $t$ or not. 
With this novel slot-based message passing in a layer, \ours are able to maintain distinguishable representations of a node in a finer granularity. 

After $L$ layers of  slot-based message passing in \ours, every node $v$ has $|\Phi|$ slot representations as shown in Figure \ref{fig:arch}. 
For downstream tasks, we develop a slot  attention technique in Section \ref{sec:slot attention} to learn slot importance scores to get final representations, which are then fed into downstream tasks. 
Algorithm \ref{alg:main} shows the pseudo code of \ours.

\subsection{Node Type  Slot Initialization}\label{sec:init}

We create $|\Phi|$ type-specific slots for every node.  
Then given a node $v$ with type $\phi(v)$, its slot $t$ $\bh_{\cv}^{(0),t}$ is initialized by Eq. \ref{eq:slotInit}.  
Specifically,  if   slot $t$ is not type $\phi(v)$, then $\bh_{\cv}^{(0),t}$ is set to zero initially; otherwise, $\bh_{\cv}^{(0),t}$ is initialized by the feature vector of $v$ with a $t$ type-specific linear transformation $\bW_{t}^{(0)}$. 
Since the feature vectors of different types of nodes can be in  different dimensions,   $\bW_{t}^{(0)}$ is a  pre-processor  to map heterogeneous node features to the same dimension $d_1$. 
\begin{equation}\label{eq:slotInit}
    \begin{aligned}
    \small
\bh_{\cv}^{(0),t}=&\begin{cases}\bW_{t}^{(0)}\bx_{\cv}\in\real^{d_1}&\text{ if }t=\phi(v),\\
        \mathbf{0}\in\real^{d_1} &\text{ if }t\neq\phi(v),
        \end{cases}
    \end{aligned}
\end{equation}
where $\bW_{t}^{(0)}\in\real^{d_1\times d_0^{\phi(\cv)}}$   is a $t$-type transformation matrix.

\subsection{Slot-based Message Passing Layer}\label{sec:messagePassing}
 
In this section, we present the   slot-based message passing layer of \ours with slot-specific transformations and a new attention mechanism.

In the $l$-th layer of \ours, all $t$-type slots of all nodes in $\mathcal{G}$ maintain  type-specific (or slot-specific) transformations $\bW_{t}^{(l)}$.
Given a node $v$ with   slot $t$ representation $\bh_{\cv}^{(l-1),t}$ from the $(l-1)$-th layer, in current $l$-th layer, the representation is transformed to $\hat{\bh}_{\cv}^{(l),t}$ by $\bW_{t}^{(l)}$  in Eq. \ref{eq:trans}. Note that the transformation is within the feature space of node type $t$, and does not affect the feature spaces of other types.
\begin{equation}\label{eq:trans} 
    \small
    \hat{\bh}_{\cv}^{(l),t}=\bW_{t}^{(l)}\bh_{\cv}^{(l-1),t},
\end{equation}
where $\bW_{t}^{(l)}\in\real^{d_{l}\times d_{l-1}}$ is a slot-specific transformation.

For instance,  in Figure \ref{fig:arch}, in the first layer, the transformation matrix $\bW_{1}^{(1)}$ for slot $1$ (in red) is applied to the slots 1 of all nodes. Even if the node is not in type $t=1$, its   slot 1 will be transformed by   $\bW_{1}^{(1)}$. 
Observe that, if a slot  is empty, its message will still be zero vector after transformation. 
 
In the $l$-th layer of \ours ($0<l\leq L$),  for every slot $t$ of every node $\cv\in\cV$ ($t\in \Phi$), we perform the above type-specified linear transformation to get slot messages $\hat{\bh}_{\cv}^{(l),t}$, which will be aggregated in the $l$-th layer via a  generic slot-based message aggregation function $aggr$. 
In particular,   given a node $v$, its  slot $t$ receives  messages $\hat{\bh}_{\cu}^{(l),t}$ from its neighbors $u\in N(v)$, which are then aggregated to update its slot $t$ to be $\bh_{\cv}^{(l),t}$,
\begin{equation}\label{eq:aggr}
\small
        \bh_{\cv}^{(l),t}=aggr\left(\{  \hat{\bh}_{\cu}^{(l),t}  \}_{\cu\in N(\cv)}\right), \forall t\in\Phi,
\end{equation}
where $\bh_{\cv}^{(l),t}\in\real^{d_{l}}$, $\bh_{\cv}^{(l),t}\in\real^{d_{l}}$. 

\begin{figure}[!t]
    \centering
    \includegraphics[width=0.66\columnwidth]{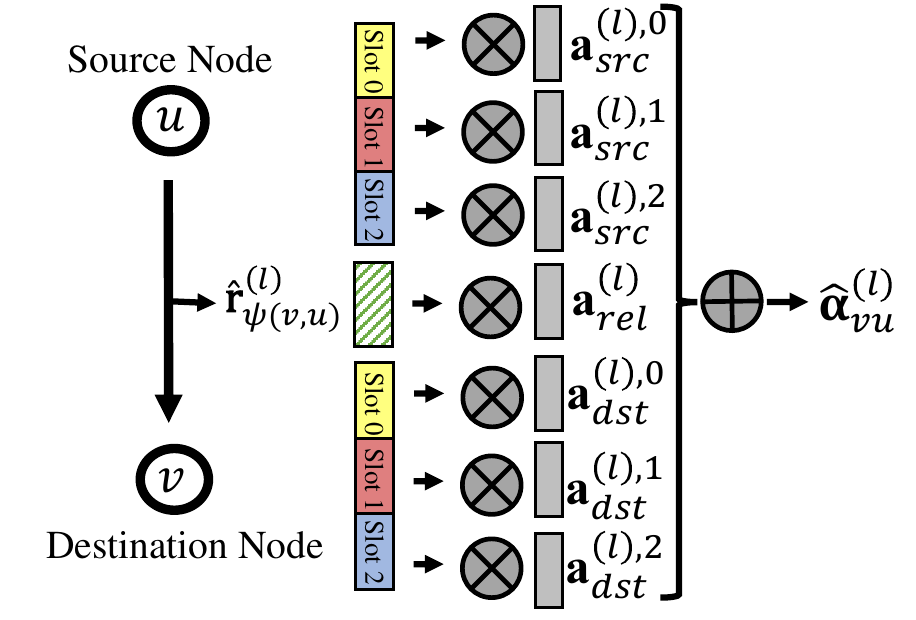}
    \caption{Attention mechanism in a message passing layer.}\label{fig:attention1}
\end{figure}

Observe that in the propagation and aggregation process above, \ours uses the whole graph topology to propagate slot messages. 
Moreover, there can be multiple options for the $aggr$ function, \eg mean and max \citep{hamilton2017inductive}.  
Instead of using these vanilla aggregation functions, we develop  an attention mechanism to learn the aggregation weights $\alpha^{(l)}_{\cv\cu}$ in Eq. \ref{eq:attn_aggr}, where the aggregation result is then passed via non-linear ReLu activation $\sigma$.
\begin{equation}\label{eq:attn_aggr}
        \small
        \bh_{\cv}^{(l),t}=\sigma\left(\sum_{\cu\in N(\cv)}\alpha^{(l)}_{\cv\cu}\hat{\bh}_{\cu}^{(l),t}\right), \forall t\in \Phi.
\end{equation}
Now we elaborate how to get aggregation weight $\alpha^{(l)}_{\cv\cu}$.
Intuitively, node $u$ passes all its slot messages to $v$ for all slots $t\in\Phi$. Thus, weight $\alpha^{(l)}_{\cv\cu}$ should be computed by considering  all the slot messages from $u$ to $v$, as well as edge types, instead of only using a specific slot $t$. In other words, all slot messages from node $u$ to $v$ share the same aggregation weight. 
Specifically,we develop a self-attention technique with slot message encoders and edge-type encoders to compute $\alpha^{(l)}_{\cv\cu}\in\real$, as illustrated in Figure \ref{fig:attention1}.
In the $l$-th layer, all nodes serving as the source of messages (\eg $u$ in Figure \ref{fig:attention1}) share a slot-specific attention vector 
$\ba^{(l),t}_{src}\in\real^{d_l}$ for every $t\in\Phi$. 
Similarly, all nodes that serve as the destination to receive messages (\eg $v$ in Figure \ref{fig:attention1}) share a slot-specific attention vector 
$\ba^{(l),t}_{dst}\in\real^{d_l}$ for every $t\in\Phi$.
Further, let $\ba^{(l)}_{rel}\in\real^{d_e}$ be an attention vector for edge types.
Then the overall  attention vector $\ba^{(l)}$ is obtained by  
\begin{equation}\label{eq:attn_vector}
\small        \ba^{(l)}=\left(\bigparallel_{t\in\Phi}\ba^{(l),t}_{src}\right) \Big| \left(\bigparallel_{t\in\Phi}\ba^{(l),t}_{dst}\right) \Big| \ba^{(l)}_{rel}.
\end{equation}
To get the edge type embedding of $l$-th layer, $\hat{\br}^{(l)}_{\psi(\cv,\cu)}$, we use a similar idea in \cite{lv2021simple}. 
Let  $\br_{\psi(\cv,\cu)}\in \real^{d_e}$ be a learnable   representation of edge type $\psi(\cv,\cu)$, and it is randomly initialized and shared across all layers for all edges with type $\psi(\cv,\cu)$.
It is transformed to get the $l$-th layer representation ${\textstyle\hat{\br}^{(l)}_{\psi(\cv,\cu)}=\bW_{rel}^{(l)}\br_{\psi(\cv,\cu)}}$, where ${\textstyle\bW_{rel}^{(l)}\in \real^{d_e\times d_e}}$ is a learnable matrix.
Then  we  compute the attention score $\alpha_{\cv\cu}^{(l)}$ by  Eq. \ref{eq:attn_score} and  \ref{eq:attn_weight}.
In Eq. \ref{eq:attn_score}, we first get an intermediate   $\hat{\alpha}_{\cv\cu}^{(l)}$ by inner product between the corresponding attention vectors and representations, including source node's slot representations ${\textstyle\hat{\bh}_{\cu}^{(l),t}}$, destination node's slot representations ${\textstyle\hat{\bh}_{\cv}^{(l),t}}$, and edge type representations ${\textstyle\hat{\br}^{(l)}_{\psi(\cv,\cu)}}$.
Then we use softmax normalization and LeakyReLU to get ${\textstyle\alpha^{(l)}_{\cv\cu}}$ in Eq. \ref{eq:attn_weight}.
\begin{equation}\label{eq:attn_score}
\small
        \hat{\alpha}_{\cv\cu}^{(l)}=\sum_{t\in\Phi } \innerPro{\ba^{(l),t}_{dst}}
    {\hat{\bh}_{\cv}^{(l),t}}+\sum_{t\in\Phi } \innerPro{\ba^{(l),t}_{src}}
        {\hat{\bh}_{\cu}^{(l),t}}+\innerPro{\ba^{(l)}_{rel} }{\hat{\br}^{(l)}_{\psi(\cv,\cu)}} 
\end{equation}
\begin{equation}\label{eq:attn_weight}
\small
        \alpha^{(l)}_{\cv\cu}=  \frac{\exp\left(\leakyrelu\left(    
        \hat{\alpha}_{\cv\cu}^{(l)}
        \right)\right)
        }{\textstyle\sum_{\cu\in N(\cv)}\exp\left(\leakyrelu\left(\hat{\alpha}_{\cv\cu}^{(l)} \right)\right)} 
\end{equation}
After $L$ slot-based message passing layers, \ours outputs the representation $\bh^{(L),t}_{\cv}$ of every slot $t$ in every node $\cv \in \cV$.

 To provide stability to the training process, \ours employs multi-head attention mechanism with $K$ heads \cite{han,velickovic2017graph,hu2020heterogeneous},
\begin{equation}\label{eq:multiHead}
\small
    \begin{aligned}
        \hat{\bh}_{\cv,(k)}^{(l),t}=&\bW_{t,(k)}^{(l)}\bh_{\cv}^{(l-1),t}\in\real^{d_{l}},\\
\bh_{\cv}^{(l),t}=&\textstyle\bigparallel_{k=1,2,...,K}\sigma\left(\textstyle\sum_{\cu\in N(\cv)}\alpha^{(l)}_{\cv\cu,(k)}\hat{\bh}_{\cu,(k)}^{(l),t}\right),\\
    \end{aligned}
\end{equation}
where subscript $(k)$ represents the corresponding variables for the $k$-th head.

\subsection{Slot Attention}\label{sec:slot attention}

After $L$ layers, \ours returns slot representations $\bh^{(L),t}_{\cv}$ for   $|\Phi|$ slots of   nodes $\cv \in \cV$.
To handle the downstream tasks with objectives to be elaborated in Section \ref{sec:objective}, we explain how to leverage all  slot representations of  node $v$ to get its final representation $\bh_v$ by a slot attention technique.

Given the slot representations of   node $v$, $\bh^{(L),t}_{\cv},\forall t\in\Phi$, a simple way is to average them by Eq. \ref{eq:average}, which however, does not differentiate slot importance. 
\begin{equation}\label{eq:average}
\small
    \begin{aligned}
        \bh_{\cv} =\frac{1}{|\Phi|}\textstyle\sum_{t\in\Phi}\bh_{\cv}^{(L),t}.
    \end{aligned}
\end{equation}
Another way is to stack a  fully connected layer,
\begin{equation}\label{eq:last fc}
\small
    \begin{aligned}
        \bh_{\cv} =\bW_{fc}\bh_{\cv}^{(L),t},
    \end{aligned}
\end{equation}
where $\bW_{fc}\in\real^{d_{task}\times (|\Phi|d_{L})}$ is a learnable matrix.

\begin{figure}[!t]

    \centering
    \includegraphics[width=0.6\columnwidth]{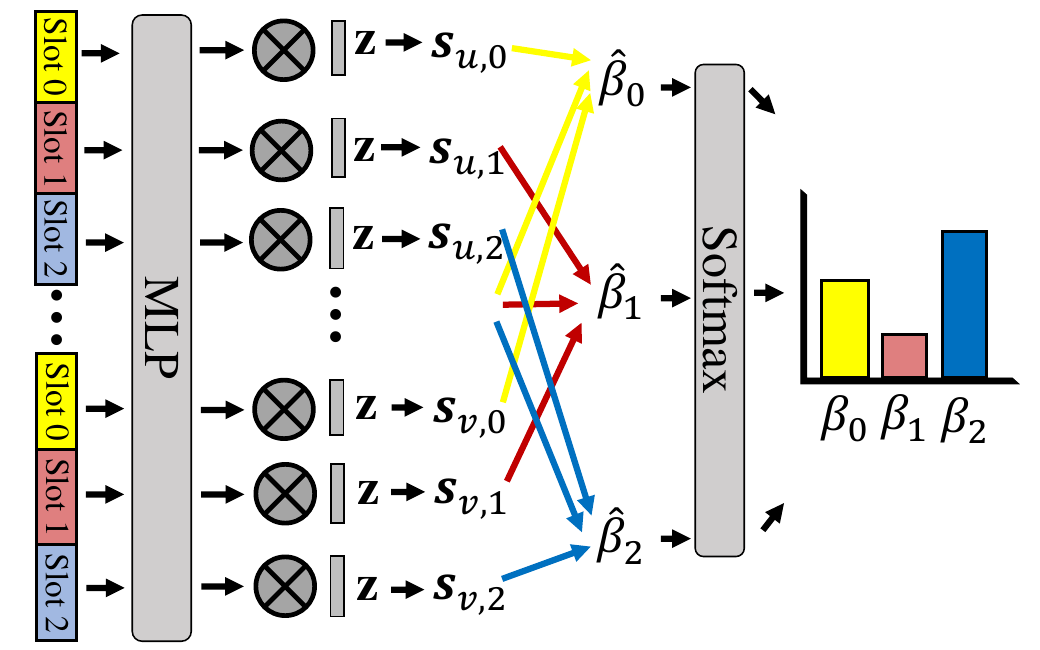}
\vspace{-2mm}
    \caption{Slot attention }\label{fig:attention2}
    \vspace{-2mm}
\end{figure}

But the way above is not explainable to demonstrate the importance of different slots in different tasks. 
Therefore, we develop a slot attention technique (illustrated in Figure \ref{fig:attention2}), which achieves better performance than the two ways above as validated in experiments.

Specifically, in   slot attention, we first apply a one-layer MLP  to transform every slot representation ${\textstyle\bh_v^{(L),t}}$ of every node $v$ of the  $L$-th layer   to vector $\mathbf{s}_{v,t}$,
\begin{equation}\label{eq:encoded semantics}
    \small
    \begin{aligned}
        \mathbf{s}_{\cv,t}=\tanh\left(\bW_b \bh_{\cv}^{(L),t} +\bb_b \right),
    \end{aligned}
\end{equation}
where ${\small\textstyle\bW_b\in\real^{d_s\times d_{L}}}$, ${\small\textstyle\bb_b\in\real^{d_s}}$ are    weights and bias.

 $\mathbf{s}_{\cv,t}$ represents the slot semantics decoded from   slot $t$ of node $\cv$ in last $L$-th layer. 
Then, for each slot $t$, we  get its semantic importance, by computing the expected similarity between the decoded semantics $\mathbf{s}_{\cv,t}$ and a learnable attention vector $\bz\in\real^{d_{L}}$, which is shared across all slots of all nodes. For slot $t$, its  intermediate attention score (without normalization) $\hat{\beta}_{t}$ is computed as follows, which considers the $t$-th slots of all nodes in $\cV$.
Then we apply softmax to all $\hat{\beta}_{t}, \forall t\in\Phi$, to get the slot attention weight $\beta_t$,
\begin{equation}
\small
        \hat{\beta}_{t}=\frac{1}{|\gV|}\textstyle\sum_{\cv\in \gV} \innerPro{\bz}{\mathbf{s}_{\cv,t}};
        \beta_{t}=\frac{\exp\left(\hat{\beta}_{t}\right)}{\sum_{\theta\in \Phi}\exp\left(\hat{\beta}_{\theta}\right)}.
\end{equation}
Finally, we aggregate the  representations of all slots in a node $v$ to get its final representation $\bh_v$.
Let matrix $\bH$ represented all  representations $\bh_v$ of all nodes $v\in\cV$.
\begin{equation}\label{eq:semanticAverage}
\small       \bh_{\cv}=\sum_{t\in\Phi}\beta_{t}\bh_{\cv}^{(L),t}.
\end{equation}

\begin{algorithm}[!t]
\caption{\ours  }\label{alg:main}
\small
\begin{algorithmic}
\STATE \textbf{Input}: Heterogeneous graph $\gG=(\cV,\gE,\phi,\psi)$, number of heads $K$, number of layers $L$, number of epochs $T$

\STATE \textbf{Output}: The learned representations $\bH$.

\STATE Generate initial parameters for all learnable parameters. 
\FOR{\text{\rm each epoch} $i=0,1,2,...,T$}
    \STATE Perform node type based slot initialization to get $\bh_{\cv}^{(0),t}$; 
        
    \FOR{\text{\rm each layer} $l=0,1,2,...,L$}
    \FOR{\text{\rm each head} $k=0,1,2,...,K$}
    
           \STATE Construct slot messages ${\textstyle\hat{\bh}_{\cv,(k)}^{(l),t}=\bW_{t,(k)}^{(l)}\bh_{\cv}^{(l-1),t}}$; 
            
            \STATE Compute attention ${\textstyle\alpha_{\cv\cu,(k)}^{(l)}}$;
            \ENDFOR
            \STATE Aggregate the messages with attention
            \STATE $\bh_{\cv}^{(l),t}\leftarrow\bigparallel_{k=1,2,...,K}\sigma(\sum_{\cu\in N(\cv)}\alpha^{(l)}_{\cv\cu,(k)}\hat{\bh}_{\cu,(k)}^{(l),t})$; 
                    
    \ENDFOR
    \STATE Compute semantics vectors $\mathbf{s}_{\cv,t}\leftarrow\text{MLP}( \bh_{\cv}^{(L),t}  )$;

    \STATE Compute    slot attention $\beta_{t}$;
  
    \STATE Aggregate the slots ${\textstyle\bh_{\cv}\leftarrow\sum_{t\in\Phi}\beta_{t}\bh_{\cv}^{(L),t}}$; 
\ENDFOR
\end{algorithmic}
\end{algorithm}

\subsection{Downstream Objectives}\label{sec:objective}
\textbf{Node Classification.} The representation  $\bh_v$ is used as the final output prediction vector for every node $\cv\in\gV$. 
Supposing that  there are $C$ classes, $\bh_v$ is in $\mathbb{R}^{d_C}$.
Let $\bH_{\cv,c}$ be the probability that node  $\cv$ is predicted to have class label $c$.

Denote $\bY$  as the groundtruth label matrix whose entries  $\bY_{\cv,c}=1$ if node $\cv$ has label $c$;  otherwise $\bY_{\cv,c}=0$.
A cross-entropy loss function is used  to train the model over labeled training nodes $\small\gL_{train}\subset \gV:   \loss_{NC}(\bH,\gG)= -\frac{1}{|\gL_{train}|}\sum_{\cv\in\gL_{train}} \sum_{c\in[C]}\bY_{\cv,c}\log\left(\bH_{\cv,c}\right)$.

\textbf{Link Prediction.} 
Link prediction is   binary classification on node pairs.
Positive pairs $\gE_{pos}\subset \gV\times\gV$ are the edges actually in $\gE$, and negative pairs $\gE_{neg}\subset \gV\times\gV$ are non-existing ones. $\gE_{pos}$ and $\gE_{neg}$ together are training samples. The model is expected to distinguish node pairs in test set whether it should be  positive or negative. 
 Given two nodes $v,u$ with representations $\bH_v$ and $\bH_u$ respectively, a common method is to decode them to get a similarity score, which is then fed into a binary cross-entropy loss:
$\small
        \loss_{LP}(\bH,\gG)=-\sum_{(\cv,\cu)\in\gE_{pos}}\log\left[Decoder\left(\bH_{\cv},\bH_{\cu}\right)\right]
        -\sum_{(\cv,\cu)\in\gE_{neg}}\log\left[1-Decoder\left(\bH_{\cv},\bH_{\cu}\right)\right]
    $. Two decoders are considered: dot product and DistMult \cite{yang2014embedding}, with details  in Appendix \ref{app:decoders}.

\subsection{Analysis}
We conduct theoretical analysis to motivate the design of slots for different node types to learn different semantics, which mitigates the semantic mixing issue. 
We adopt the row-normalized Laplacian model in \citep{gcn} for analysis, and the empirical property of attention models are similar to spectra-based graph convolution methods with row-normalization \citep{oversmoothingEmpirical}.

Let $A$ be the adjacency matrix and $D$ be the degree matrix of     heterogeneous graph $\mathcal{G}$, and define $L_{rw}=D^{-1}(D-A)$. The spectral graph convolution operator $G$ is  $G=I-\alpha L_{rw}$, where $\alpha\in(0,1]$.
Suppose that  $\mathcal{G}$ has $k$ maximal connected components (CC) $\{CC_i\}_{i=1}^{k}$.  
Let $cc(j)$ denote the CC where node ${v}_j$ is, and $|cc(j)|$ be the size of the CC.
Considering the slot initialization in Section~\ref{sec:init}, let $\mathbf{X}^{(t)}\in \mathbb{R}^{n\times d^t_0}$ be the initial feature matrix of slots $t$ of all $n$ nodes   in $\mathcal{G}$, and the $i$-th row vector  $\mathbf{x}_i^{(t)}$ be the initial feature vector of   slot $t$ of node $v_i$.
Only nodes in type $t$ have nonzero $\mathbf{x}_i^{(t)}$ in  $\mathbf{X}^{(t)}$; otherwise,  $\mathbf{x}_i^{(t)}=0$.
With self-loop added to every node, there is no bipartite component in the graph. 
Then inspired by~\citep{li2018deeper}, we derive  Theorem~\ref{thm:main}, with proof in Appendix \ref{sec:proof of main}. 
\begin{theorem}\label{thm:main}
Given a heterogeneous graph $\mathcal{G}$, for the $t$-slot feature matrix $\mathbf{X}^{(t)}\in \mathbb{R}^{n\times  d^t_0}$ in which only nodes of type $t$ have non-zero features, if the graph has no bipartite components, after infinite number of  convolution operations $G$, the slot $t$ representation of node $v_i$ is
\begin{equation*}
\small
[\lim_{\ell\to+\infty}G^\ell \mathbf{X}^{(t)}]_i
    =\frac{1}{|cc(i)|} \sum_{ \forall v_j\in cc(i), \phi(\cv_{j})=t}
    \mathbf{x}^{(t)}_j.
\end{equation*}
\end{theorem}
As proved in the theorem, after infinite steps, the representation   of the slot $t$ in node $v_i$ converges to the average feature vector of all the nodes in the same type $t$ in $cc(i)$. 
 Since the semantics of the nodes in a different node type $t'$ with initial features $\mathbf{X}^{(t')}$, are usually very different from $\mathbf{X}^{(t)}$,
the representations for slots $t'$ will apparently  converge to a different state.
Theorem~\ref{thm:main} proves that the slots of the nodes within the \textit{same} connected component $cc(i)$
 converge to \textit{different} states $\frac{1}{|cc(i)|}\sum_{\forall v_j \in cc(i),\phi(v_j)=t}\mathbf{x}_j^{(t)}$ that are relevant to node type $t$
 feature spaces, which explains the power of SlotGAT to mitigate the semantic mixing issue. On the other hand,~\citep{li2018deeper} prove that all node representations in a homogeneous connected component converge to the same state.
\ours performs $L$ layers (${\textstyle\ll+\infty})$, and thus the slot representations output by \ours, $\bh^{(L),t}_{\cv_i}$ are semantically different from $\bh^{(L),t'}_{\cv_i}$ for slots $t$ and $t'$ of node $v_i$ in $\mathcal{G}$. 
We further provide visualization results of slot representations in Section \ref{sec:ablation} to   validate that the slot-based message passing in \ours captures rich semantics.

\textbf{Complexity.} We then provide the complexity analysis of \ours, by following a similar fashion as \citep{velickovic2017graph}. 
Let $m$, $n$ and $|\Phi|$ 
 be the number of nodes, edges, and node types, respectively, and assume that all representations in all layers have the same dimension $d$. In one \ours layer (Section \ref{sec:messagePassing}), (i) feature transformation in Eq.~\ref{eq:trans} needs $O(nd^2|\Phi|)$;  (ii) attention computation in Eq.~\ref{eq:attn_score} and Eq.~\ref{eq:attn_weight} needs $O(md|\Phi|)$. Hence, the complexity of a single layer is $O(nd^2|\Phi|+md|\Phi|)$, which is on par with baseline methods if factor $|\Phi|$ is regarded as a constant. According to~\citep{velickovic2017graph}, applying $K$ heads multiplies the storage and parameters, while the computation of individual heads is fully independent and parallelized. The computation of Eq.~\ref{eq:slotInit} in Section~\ref{sec:init} is only performed once for all $n$ nodes before the $L$-layer message passing, and it needs time $O(ndd_0|\Phi|)$, which is $O(nd^2|\Phi|)$ if the input feature dimension $d_0$ is $d$ as well. The slot attention in Section~\ref{sec:slot attention} is also computed only once after message passing, and its complexity is $O(ndd_s|\Phi|)$, which is $O(nd^2|\Phi|)$ if the embedding dimension of slot attention $d_s$ is also $d$. The number of layers $L$ is usually small as a constant, e.g., 3. Hence, the complexity of \ours is $O(nd^2|\Phi|+md|\Phi|)$.

\begin{table}[!t]
\caption{Statistics of Benchmark Datasets.}\label{tab:benchmarks}
\small
\setlength{\tabcolsep}{0.6mm}
\resizebox{0.98\columnwidth}{!}{
\begin{tabular}{crcrccc}
\toprule
 \makecell[c]{\emph{Node}\\\emph{Classification}} & \#Nodes & \makecell[c]{\#Node\\Types} & \#Edges & \makecell[c]{\#Edge\\Types}& \makecell[c]{Target}& \#Classes\\ 
 \midrule
DBLP          & 26,128               & 4                         & 239,566              & 6                         & author                                                                             & 4                            \\
IMDB          & 21,420               & 4                         & 86,642               & 6                         & movie                                                                              & 5                            \\
ACM           & 10,942               & 4                         & 547,872              & 8                         & paper                                                                              & 3                            \\
Freebase      & 180,098              & 8                         & 1,057,688            & 36                        & book                                                                               & 7                            \\
PubMed\_NC      & 63,109              & 4                         & 244,986             & 10                       & disease                                                                               & 8                            \\
\midrule
\emph{Link Prediction}& & & & & \multicolumn{2}{c}{\makecell[c]{Target}}\\
\midrule
LastFM        & 20,612               & 3                         & 141,521              & 3                         & \multicolumn{2}{c}{user-artist}                           \\
PubMed\_LP        & 63,109               & 4                         & 244,986              & 10                        & \multicolumn{2}{c}{disease-disease} \\
\midrule
\end{tabular}
\vspace{-3mm}
}
\end{table}

\begin{table*}[!t]
\caption{Node classification results with mean and standard deviation of  Macro-F1/Micro-F1. Vacant positions (“-”) mean out of memory.
Best is in bold, and runner up is underlined.}
\label{tab:NCtotal}
\centering
\footnotesize


\setlength{\tabcolsep}{4pt}

\resizebox{1\textwidth}{!}{%
\begin{tabular}{ccccccccccc}
\hline
 &
  \multicolumn{2}{c}{DBLP} &
  \multicolumn{2}{c}{IMDB} &
  \multicolumn{2}{c}{ACM} &
  \multicolumn{2}{c}{PubMed\_NC} &
  \multicolumn{2}{c}{Freebase} \\ \hline
  &
  Macro-F1&Micro-F1 &
  Macro-F1&Micro-F1 &
  Macro-F1&Micro-F1 &
  Macro-F1&Micro-F1 &
  Macro-F1&Micro-F1  \\ \hline
RGCN &
  91.52±0.50&92.07±0.50 &
  58.85±0.26&62.05±0.15 &
  91.55±0.74&91.41±0.75 &
  18.02±1.98&20.46±2.39 &
  46.78±0.77&58.33±1.57  \\ 
HAN &
  91.67±0.49&92.05±0.62 &
  57.74±0.96&64.63±0.58 &
  90.89±0.43&90.79±0.43 &
   15.43±2.41&24.88±2.27 &
  21.31±1.68&54.77±1.40 
   \\ {
DisenHAN} &
  93.66±0.39&94.18±0.36 &
  63.40±0.49&\underline{67.48±0.45} &
  92.52±0.33&92.45±0.33 &
  41.71±4.43&50.93±4.25&
  -&- \\ 
GTN &
  93.52±0.55&93.97±0.54 &
  60.47±0.98&65.14±0.45 &
  91.31±0.70&91.20±0.71 &
  - & - &
  - & - \\
RSHN &
  93.34±0.58&93.81±0.55 &
  59.85±3.21&64.22±1.03 &
  90.50±1.51&90.32±1.54 &
  - & - &
  - &- \\
HetGNN &
  91.76±0.43&92.33±0.41 &
  48.25±0.67&51.16±0.65 &
  85.91±0.25&86.05±0.25 &
  21.86±3.21&29.93±3.51 &
  - &- \\ 
MAGNN &
  93.28±0.51&93.76±0.45 &
  56.49±3.20&64.67±1.67 &
  90.88±0.64&90.77±0.65 &
  - & -&
  - & - \\ 
HetSANN &
  78.55±2.42&80.56±1.50&
  49.47±1.21&57.68±0.44&
  90.02±0.35&89.91±0.37&
  - & -&
  - & -\\ 
HGT &
  93.01±0.23&93.49±0.25 &
  63.00±1.19&67.20±0.57 &
  91.12±0.76&91.00±0.76 &
  \underline{47.50±6.34}&\underline{51.86±4.85} &
  29.28±2.52&60.51±1.16  \\ \hline
GCN &
  90.84±0.32&91.47±0.34 &
  57.88±1.18&64.82±0.64 &
  92.17±0.24&92.12±0.23 &
  9.84±1.69&21.16±2.00  &
  27.84±3.13&60.23±0.92  \\ 
GAT &
  93.83±0.27&93.39±0.30 &
  58.94±1.35&64.86±0.43 &
  92.26±0.94&92.19±0.93 &
  24.89±8.47&34.65±5.71  &
  40.74±2.58&65.26±0.80  \\ \hline
simpleHGN &
  94.01±0.27&94.46±0.20 &
  \underline{63.53}±1.36&67.36±0.57 &
  \underline{93.42±0.44}&\underline{93.35±0.45} &
  42.93±4.01&49.26±3.32 &
  \underline{47.72±1.48}&\underline{66.29±0.45}  \\ 
space4HGNN &
  \underline{94.24±0.42}&\underline{94.63±0.40} &
  61.57±1.19&63.96±0.43 &
  92.50±0.14&92.38±0.10 &
  45.53±4.64&49.76±3.92 &
  41.37±4.49&65.66±4.94 \\ \hline
\ours &
  \textbf{94.95±0.20}&\textbf{95.31±0.19} &
  \textbf{64.05±0.60}&\textbf{68.54±0.33} &
  \textbf{93.99±0.23}&\textbf{94.06±0.22} &
  \textbf{47.79±3.56}&\textbf{53.25±3.40} &
  \textbf{49.68±1.97}&\textbf{66.83±0.30} \\ \hline
\end{tabular}
}

\vspace{-2mm}
\end{table*}

\section{Experiments}\label{sec:exp}

\subsection{Experiment Settings}
\textbf{Datasets.}
Table \ref{tab:benchmarks} reports the statistics of benchmark datasets widely used in   \cite{lv2021simple,han,space4HGNN,hetgnn,gtn,hne}. 
The datasets cover various domains, including  academic graphs (\eg DBLP, ACM),  information  graphs (\eg IMDB, LastFM, Freebase), and medical biological graph (\eg PubMed). 
For node classification, each dataset contains a target node type, and all nodes with the target type are the nodes for classification \cite{lv2021simple}.
PubMed\_LP is the same as PubMed\_NC, but for link prediction. 
Each link prediction dataset has a target edge type, and edges in the target edge type are the target for prediction.
The descriptions of all  datasets are in Appendix \ref{app:data}.

\textbf{Baselines.}
We compare with HAN~\citep{han}, {DisenHAN~\citep{disenHAN}}, GTN~\citep{gtn},
MAGNN~\citep{magnn}, HetGNN~\citep{hetgnn}, HGT~\citep{hu2020heterogeneous}, RGCN~\citep{schlichtkrull2018modeling}, RSHN\citep{rshn}, HetSANN~\citep{hong2020attention}, simpleHGN~\citep{lv2021simple}, and Space4HGNN~\citep{space4HGNN}.
We also compare with GCN~\citep{gcn} and GAT~\citep{velickovic2017graph}.

\textbf{Evaluation Settings.}
For node classification, following \cite{lv2021simple}, we split labeled training set into training and validation  with   ratio $80\%:20\%$, while the testing data are fixed with detailed numbers in Appendix \ref{app:datasplit} Table \ref{tab:benchmark split}.
For link prediction, we adopt ratio $81\%:9\%:10\%$  to divide the  edges into training, validation, and testing. 
For each dataset, we repeat experiments on 5 random   splits
and then report the average and standard deviation. 
For node classification, we report the averaged Macro-F1 and Micro-F1 scores. 
We also conduct paired t-tests~\citep{ttest,klicpera2018predict} to evaluate the statistical significance of \ours.
For link prediction, we report MRR (mean reciprocal rank) and ROC-AUC.
The  evaluation metrics are summarized in Appendix~\ref{sec:computation of metrics}. 
Link prediction needs negative samples (\ie non-existence edges) for test, which are generated within 2-hop neighbors of a node \citep{lv2021simple}.

\textbf{Hyper-parameter Search Space.}
We search learning rate within $\{1,5\}\times\{1e^{-5},1e^{-4},1e^{-3},1e^{-2}\}$, weight decay rate within  $\{1,5\}\times\{1e^{-5},1e^{-4},1e^{-3}\}$, dropout rate for features within $\{0.2,0.5,0.8,0.9\}$, dropout rate for connections within $\{0,0.2,0.5,0.8,0.9\}$, and number of hidden layers $L$ within $\{2,3,4,5,6\}$. We use the same dimension of hidden embeddings  across all layers $d_l$ within $\{32,64,128\}$.
We search the number of epochs within the range of $\{40,300,1000\}$ with early stopping patience $40$, 
and dimension $d_s$ of slot attention vector within the range of $\{3,8,32,64\}$. 
Following \citep{lv2021simple}, for input feature type, we use feat = 0 to denote the use of  all given
features, feat = 1 to denote using only target node features (zero vector for others), and
feat = 2 to denote all nodes with one-hot features. 
For node classification, we use  feat 1 and set the number of attention heads $K$ to be 8. For link prediction, we use feat 2 and set  $K$ to be 2. 
Link prediction has two decoders, dot product and DistMult, which are also searchable. 
The search space and strategies  of all baselines follow~\citep{lv2021simple,space4HGNN}.
The searched hyper parameters are in Appendix \ref{app:param}.

\begin{table}[!t]
\caption{Link prediction results  with mean and standard deviation of ROC-AUC/MRR.  Vacant positions (“-”) means  out of memory. }
\label{tab:LPtotal}
\centering
\small


\resizebox{0.99\columnwidth}{!}{
\begin{tabular}{ccccc}
\hline
 &
  \multicolumn{2}{c}{LastFM} &
  \multicolumn{2}{c}{PubMed\_LP} \\ \hline
  &
  ROC-AUC&MRR &
  ROC-AUC&MRR  \\ \hline
RGCN &
  57.21±0.09&77.68±0.17 &
  78.29±0.18&90.26±0.24 \\ 
  {DisenHAN}	& 57.37±0.2&76.75±0.28&73.75±1.13&85.61±2.31 \\
HetGNN &
  62.09±0.01&83.56±0.14 &
  73.63±0.01&84.00±0.04 \\ 
MAGNN &
  56.81±0.05&72.93±0.59 &
  -  &   - \\
HGT &
  54.99±0.28&74.96±1.46 &
  80.12±0.93&90.85±0.33 \\ \hline
GCN &
  59.17±0.31&79.38±0.65 &
  80.48±0.81&90.99±0.56 \\ 
GAT &
  58.56±0.66&77.04±2.11 &
  78.05±1.77&90.02±0.53 \\ \hline
simpleHGN &
  \underline{67.59±0.23}&\underline{90.81±0.32} &
  \underline{83.39±0.39}&\underline{92.07±0.26} \\ 
space4HGNN &
  66.89±0.69&90.77±0.32 &
   81.53±2.51&90.86±1.02\\  \hline
\ours &
  \textbf{70.33±0.13}&\textbf{91.72±0.50} &
  \textbf{85.39±0.28}&\textbf{92.22±0.28} \\ \hline
\end{tabular}
\vspace{-2mm}
}


\end{table}

\subsection{Evaluation Results} 
\textbf{Node Classification.} 
Table~\ref{tab:NCtotal} reports the node classification results of all methods on five datasets, with mean Macro-F1 and Micro-F1 and their standard deviation. 
The overall observation is that \ours consistently outperforms all baselines for both  Macro-F1 and Micro-F1 on all datasets, often by a significant margin, which validates the power of  \ours.
For instance, in IMDB that is a multi-label classification dataset, the Micro-F1 of \ours is 68.54\%,  while that of the best competitor is {67.48\%}; moreover, \ours has a smaller standard deviation 0.33, compared to 0.57 of the competitor. 
On the largest Freebase, \ours achieves 49.68\% Macro-F1, while the best competitor has 47.72\%.
On PubMed\_NC,  \ours has  higher Micro-F1 53.25\% compared with 51.86\% of HGT.
On DBLP and ACM, where all methods are with relatively high performance, \ours still achieves the best performance.
The superior performance of \ours on node classification  indicates that \ours with slot-based message passing mechanism is   able to learn richer representations to preserve the node heterogeneity in various  heterogeneous graphs.

\textbf{Link Prediction.}
Table \ref{tab:LPtotal} reports the link prediction results on LastFM and PubMed\_LP, with mean ROC-AUC and MRR   and their standard deviations. 
\ours achieves the best performance on both datasets over both metrics. 
For instance, 
on LastFM, \ours is with ROC-AUC 70.33\%, 2.74\% higher than simpleHGN, the best competitor, and also the MRR of \ours is 91.72\%, while that of simpleHGN is 90.81\%. 
On PubMed\_LP, \ours is with ROC-AUC 85.39\%, 2\% higher than simpleHGN. 
The performance of \ours for link prediction   again reveals the effectiveness of the proposed slot-based message passing mechanism and attention techniques to avoid semantic mixing issue and preserve pairwise node relationships for link prediction.

\subsection{Model Analysis}\label{sec:ablation}

\begin{table}[!t]
\centering
\caption{Statistical Significance Test (Node Classification)}\label{tbl:significance test}
\small
\setlength{\tabcolsep}{4pt}
\resizebox{0.96\columnwidth}{!}{
\begin{tabular}{llllll} 
\hline
 & DBLP & IMDB & ACM & PubMed\_NC & Freebase \\
\hline
micro-F1 & 0.000244 & 0.000111 & 0.00713 & 0.00102 & 0.0421 \\ 
\hline
macro-F1 & 0.000209 & 0.160 & 0.00783 & 0.000192 & 0.156 \\ 
\hline
\end{tabular}
}
\end{table}

\begin{table}[!t]
\centering
\caption{{Statistical Significance Test (Link Prediction)}}\label{tbl:significance test LP}
\small
\setlength{\tabcolsep}{4pt}
\resizebox{0.48\columnwidth}{!}{
\begin{tabular}{lll} 
\hline
 &LastFM &	PubMed\_LP \\
\hline
ROC-AUC	&0.0252&	0.00801\\
\hline
MRR&	0.0445&	0.0261\\
\hline
\end{tabular}
}
\vspace{-2mm}
\end{table}

\begin{figure}[!t]
    \centering
    \includegraphics[width=.36\textwidth]{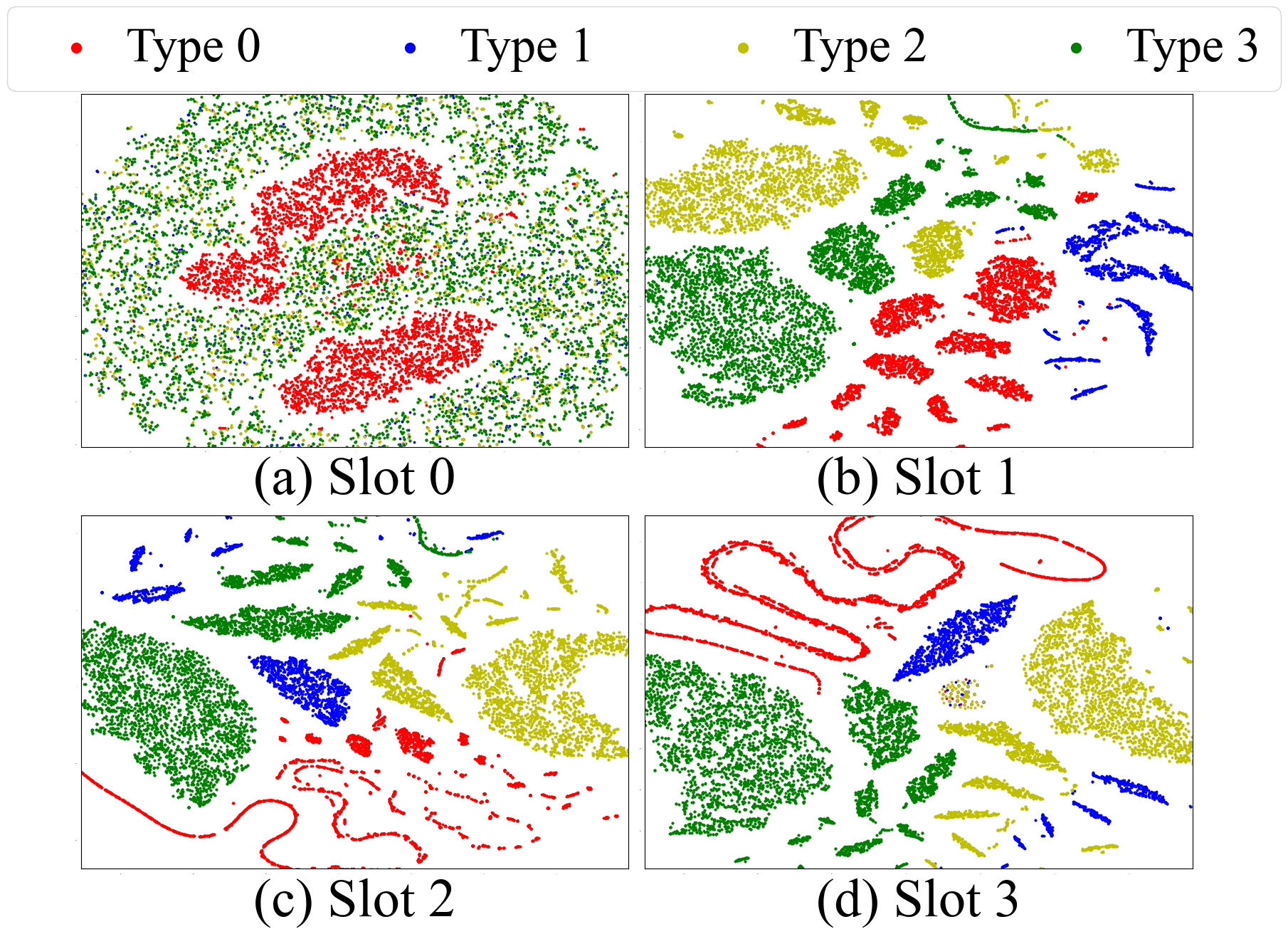}
        \vspace{-2mm}
    \caption{Tsne visualization on all slot representations of $2$-nd \ours layer on IMDB}
    \label{fig:tsneIMDB}
    \vspace{-2mm}
\end{figure}

\textbf{Slot Visualization.}
We adopt Tsne~\cite{tsne} to visualize the slot representations of \ours.
Figure \ref{fig:tsneIMDB} reports the slot visualization of the 2-nd \ours layer on IMDB. 
Figure \ref{fig:tsneIMDB}(a) is the visualization of the 0-th slots 
(in type-0 feature space)
of all nodes with colors representing their own types.
Figure \ref{fig:tsneIMDB}(b), (c), (d) are the visualization of 1-st,  
2-nd,
and 3-rd slots
of all nodes respectively. 
First, observe that the 4 slots of a node (for all nodes in a graph) learn radically different representations, as the visualizations in the four figures exhibit different patterns. 
For instance, in Figure \ref{fig:tsneIMDB}(a), 0-th slots   can distinguish nodes in type 0 (red) from others, but cannot distinguish nodes of types 1, 2, 3. 
On the other hand, in Figure \ref{fig:tsneIMDB}(b), 1-st slots can distinguish nodes of all types in 0, 1, 2, 3.
Similar observations can be made in Figures \ref{fig:tsneIMDB}(c) (d).
The visualization in Figure \ref{fig:tsneIMDB} validates that \ours is able to learn richer semantics in a finer granularity and also preserve the semantic differences of different node-type feature spaces, which explains the superior performance of \ours.
Additional visualization on DBLP is provided in Appendix \ref{app:visual}.

\begin{table*}[!t]
\caption{Ablation for Slot Attention (Macro-F1/Micro-F1) on node classification. Best is in bold, and runner up is underlined.}
\label{tab:ablateAttention}
\centering
\small
\setlength{\tabcolsep}{3pt}
\resizebox{0.999\textwidth}{!}{
\begin{tabular}{c ccccc ccccc}
\hline
 &
  \multicolumn{2}{c}{DBLP} &
  \multicolumn{2}{c}{IMDB} &
  \multicolumn{2}{c}{ACM} &
  \multicolumn{2}{c}{PubMed\_NC} &
  \multicolumn{2}{c}{Freebase}  \\ \hline
  &
  Macro-F1&Micro-F1 &
  Macro-F1&Micro-F1 &
  Macro-F1&Micro-F1 &
  Macro-F1&Micro-F1 &
  Macro-F1&Micro-F1  \\ \hline
\ours &
  \textbf{94.95±0.20}&\textbf{95.31±0.19} &
 \textbf{ 64.05±0.60}&\textbf{68.54±0.33} &
  \textbf{93.99±0.23}&\textbf{94.06±0.22} &
  \underline{47.79±3.56}&\underline{53.25±3.40} &
  \underline{49.68±1.97}&\underline{66.83±0.30} \\ \hline
\ours(average) &
  94.52±0.31&94.93±0.25 &
  62.71±1.10&67.56±0.42 &
  93.43±0.50&93.35±0.43 &
  47.18±4.16&52.61±3.23 &
   47.90±1.61&67.01±0.90  \\ \hline
\ours(target slot) &
  87.12±8.44&87.73±7.91 &
  39.35±18.9&48.91±11.2 &
  93.63±1.03&93.50±1.15 &
  44.60±4.14&50.54±4.65 &
   48.77±2.86&66.38±0.50 \\ \hline
\ours(last fc) &
 \underline{94.72±0.22}&\underline{95.11±0.21} &
  \underline{63.15±0.53}&\underline{68.16±0.31}  &
  \underline{93.91±0.35}&\underline{93.99±0.35} &
  \textbf{48.24±5.37}&\textbf{53.72±2.88} &
   \textbf{51.63±0.75}&\textbf{66.68±0.42}  \\ \hline
\end{tabular}
}
\end{table*}

\begin{table}[!t]
\centering
\caption{Statistical Significance Test  between SlotGAT and its variants w.r.t. Micro-F1.}
\label{tab:t-test-micro}
\small
\setlength{\tabcolsep}{3pt}
\resizebox{0.999\columnwidth}{!}{
\begin{tabular}{ c c c c c c }
\hline
 & DBLP & IMDB & ACM & PubMed & Freebase \\ \hline
\ours(average) & 0.0090 & 0.0362 & 0.0379 & 0.0226 & 0.3054 \\ \hline
\ours(target slot) & 0.0129 & 0.1014 & 0.0300 & 0.0248 & 0.1988 \\ \hline
\ours(last fc) & 0.0832 & 0.6714 & 0.2840 & 0.1588 & 0.2571 \\ \hline
\end{tabular}
}
\end{table}

\begin{table}[!t]
\centering
\caption{Statistical Significance Test between SlotGAT and its variants w.r.t. Macro-F1.}
\label{tab:t-test-macro}
\small
\setlength{\tabcolsep}{3pt}
\resizebox{0.999\columnwidth}{!}{
\begin{tabular}{cccccc}
\hline
 & DBLP & IMDB & ACM & PubMed & Freebase \\ \hline
\ours(average) & 0.0042 & 0.0325 & 0.0359 & 0.1833 & 0.2611 \\ \hline
\ours(target slot) & 0.0304 & 0.1074 & 0.0372 & 0.1984 & 0.1160 \\ \hline
\ours(last fc) & 0.0420 & 0.6952 & 0.2748 & 0.9704 & 0.2114 \\ \hline
\end{tabular}
}
\end{table}

\textbf{Statistical Significance Test.}
In Table \ref{tbl:significance test}, we report the \textit{p-values} of  paired t-tests  between \ours  and the best baseline simpleHGN, \textit{w.r.t.}, micro-F1 and macro-F1 for node classification. {In Table \ref{tbl:significance test LP} we report the {p-values} on link prediction \textit{w.r.t.} ROC-AUC and MRR.} P-value below 0.05 means statistical significance. Observe that most p-values are below $0.05$ except two p-values, indicating the improvement of \ours over simpleHGN is statistically significant and robust.

\textbf{Slot Attention Ablation.}
We ablate the slot attention technique in Section \ref{sec:slot attention} that integrates all slots of a node, and compare with other ways, including average (Eq. \ref{eq:average}), non-biased fully connected layer (last fc) in Eq. \ref{eq:last fc}, and target slot only (target): $\small
        \bh_{\cv}^{(L)}=\frac{1}{|\Phi_{target}|}\sum_{t\in\Phi_{target}}\bh_{\cv}^{(L),t}\in\real^{d_{L}}.$
The results are reported in  Table \ref{tab:ablateAttention}. 
Observe that \ours with slot attention  is the best on DBLP, IMDB, and ACM, and is top-2 on PubMed\_NC and Freebase.
The results validate the effectiveness of slot attention. 
Moreover, \ours(average) and \ours(target) are not with high performance, indicating that blindly averaging slot representations or only using the target slot   are less effective. 
{Moreover, We provide the paired t-test results between SlotGAT and its variants  \textit{w.r.t.} Micro-F1 and Macro-F1 in Table~\ref{tab:t-test-micro} and Table~\ref{tab:t-test-macro} respectively. Combining with the results in Table~\ref{tab:ablateAttention}, we have the following observations. (i) On DBLP, IMDB, ACM and PubMed\_NC for Micro-F1 (Table~\ref{tab:t-test-micro}), the improvement of SlotGAT over SlotGAT(average) and SlotGAT(target slot) is statistically significant, since 7 out of the 8 corresponding p-values are below 0.05, and similar observation can be made on Table~\ref{tab:t-test-macro} for these methods. (ii) As shown in Table~\ref{tab:ablateAttention}, SlotGAT(last fc) is comparable to SlotGAT, and thus the corresponding p-values are relatively large in the last rows of Tables~\ref{tab:t-test-micro} and~\ref{tab:t-test-macro}. Note that SlotGAT (last fc) is less explainable than SlotGAT with slot attention. (iii) On Freebase, all methods have close F1 scores in Table~\ref{tab:ablateAttention}, and thus the p-values are relatively large in the last columns of Tables~\ref{tab:t-test-micro} and~\ref{tab:t-test-macro}. These observations, combined with Table~\ref{tab:ablateAttention}, validate our statement  that SlotGAT with slot attention is the best on DBLP, IMDB, and ACM, and is top-2 on PubMed\_NC and Freebase.}

\textbf{Efficiency.}
Tables \ref{tbl:traintime} and \ref{tbl:infertime} report the training time   and inference time of \ours compared with  simpleHGN and HGT, two strong baselines.
Table \ref{tbl:memory} reports the peak memory usage.
As shown, \ours has moderate training and inference time and slightly higher memory usage to achieve the state-of-the-art effectiveness reported ahead.

\begin{table}[!t]
\centering
\caption{Training time per epoch (millisecond)} \label{tbl:traintime}
\resizebox{1\columnwidth}{!}{
\begin{tabular}{cccccc} 
\hline
 & DBLP & IMDB & ACM & PubMed\_NC & Freebase \\ 
\hline
simpleHGN & 76.21 & 75.57 & 105.44 & 170.77 & 423.32 \\ 
\hline
SlotGAT & 230.61 & 241.36 & 131.98 & 520.95 & 458.23 \\
\hline
HGT & 741.23 & 613.12 & 1097.8 & 2384.14 & 3711.24 \\ 
\hline
\end{tabular}
}
\end{table}

\begin{table}[!t]
\centering
\caption{Inference time (millisecond)}\label{tbl:infertime}
\resizebox{1\columnwidth}{!}{
\begin{tabular}{cccccc} 
\hline
 & DBLP & IMDB & ACM & PubMed\_NC & Freebase \\ 
\hline
simpleHGN & 57.1 & 54.42 & 47.16 & 70.67 & 175.34 \\ 
\hline
SlotGAT & 127.19 & 119.95 & 69.1 & 258.44 & 231.31 \\
\hline
HGT & 531.13 & 410.52 & 650.12 & 1818.42 & 2939.8 \\ 
\hline
\end{tabular}
}
\end{table}

\begin{table}[!t]
\centering
\caption{Peak GPU memory (GB)}\label{tbl:memory}
\resizebox{1\columnwidth}{!}{
\begin{tabular}{cccccc} 
\hline
 & DBLP & IMDB & ACM & PubMed\_NC & Freebase \\  
\hline
simpleHGN & 3.46 & 3.01 & 6.32 & 5.96 & 7.06 \\ 
\hline
SlotGAT & 6.36 & 5.97 & 5.5 & 13.37 & 9.32 \\
\hline
HGT & 1.34 & 1.08 & 2.02 & 2.54 & 6.75 \\
\hline
\end{tabular}
}
\end{table}

\section{Conclusion, Limitation, and Future Work}\label{sec:con}

In this paper, we identify a semantic mixing issue that potentially hampers the performance   on heterogeneous graphs. 
To alleviate the issue, we design \ours with separate message passing processes in slots, one for each node type.
In such a way, \ours conducts slot-based message passing. We also design an   attention-based aggregation mechanism over all slot representations to conduct effective slot-wise message aggregation per layer. To effectively support downstream tasks, we further develop a slot attention technique.
Extensive experiments validate the superiority of \ours. 

\ours has  factor $|\Phi|$ as a {trade-off} of efficiency for effectiveness, which is a potential limitation if many node types exist.
There are several ways to handle this in the future, \eg being selective on node types to reduce the number of slots and setting a limit on the dimension of all slots. 

\section*{Acknowledgement}
This work is supported by Hong Kong RGC ECS No. {25201221}, and National Natural Science Foundation of China No. {62202404}.
This work is also supported by a collaboration grant from {Tencent Technology (Shenzhen) Co., Ltd (P0039546)}.
This work is supported by a startup fund ({P0033898}) from Hong Kong Polytechnic University and project {P0036831}.

\clearpage


\bibliography{example_paper}
\bibliographystyle{icml2023}

\clearpage

\appendix
\section{Appendix}\label{sec:appendix}

All experiments are conducted on a machine powered by  an Intel(R) Xeon(R) E5-2603 v4 @ 1.70GHz CPU, 131GB RAM, and a Nvidia Geforce 3090 Cards with Cuda version 11.3.  
Source codes of all competitors are obtained from  respective authors.

\subsection{Dataset Descriptions} \label{app:data}

For all of the benchmark datasets, one could access them in online platform HGB\footnote{{\url{https://www.biendata.xyz/hgb/}}}. In the following, we provide the descriptions of these datasets. The detailed data statistics, including the number of nodes in every type  and the number of edges in every type, can be found at 
\url{https://github.com/scottjiao/SlotGAT_ICML23/data_statistics.txt}.
\vspace{-\topsep}
\begin{itemize}
    \item \textbf{DBLP}\footnote{\url{http://web.cs.ucla.edu/~yzsun/data/}} is a bibliography website of computer science. 
    There are 4 node types, including  authors, papers, terms, and venues, as well as  $6$ edge types.
    The edge types include  paper-term, paper-term, paper-venue, paper-author, term-paper, and venue-paper. 
    The target is to predict the class labels of authors. The classes are database, data mining, AI, and information retrieval. 
    \item \textbf{IMDB}\footnote{\url{https://www.kaggle.com/karrrimba/movie-metadatacsv}} is a website about movies. 
    There are 4 node types: movies, directors, actors, and keywords. A movie can have multiple class labels.
    The $6$ edge types include movie-director, director-movie, movie-actor, actor-movie, movie-keyword, and keyword-movie.
    There are 5 classes: action, comedy, drama, romance, and thriller. Movies are the targets to classify.   
    \item \textbf{ACM} is a citation network \citep{han}, containing  node types of authors, papers, terms, and subjects. 
    The  edge types include paper-cite-paper, paper-ref-paper, paper-author, author-paper, paper-subject, subject-paper, paper-term, and term-paper. 
    The node classification target is to classify papers into 3 classes: database, wireless communication, and data mining. 
    \item \textbf{Freebase}~\cite{bollacker2008freebase} is a large knowledge graph with 8 node types, including book, film, music, sports, people, location, organization, and business, and 36 edge types.  
    The target is to classify  books in 7 classes: scholarly work, book subject, published work, short story, magazine, newspaper, journal, and poem. 
    \item \textbf{LastFM}~\footnote{\url{https://grouplens.org/datasets/hetrec-2011/}} is an online music website. There are 3 node types: user, artist, and tag. And it has $3$ edge types: user-artist , user-user, artist-tag. The link prediction task aims to predict the edges between users and artists. 
    \item \textbf{PubMed}\footnote{\url{https://pubmed.ncbi.nlm.nih.gov}} is a biomedical literature library. We use the data constructed by~\cite{hne}. The node types  are gene, disease, chemical, and species. The $10$ edge types contain gene-and-gene, gene-causing-disease, disease-and-disease, chemical-in-gene, chemical-in-disease, chemical-and-chemical, chemical-in-species, species-with-gene, species-with-disease, and species-and-species. The target of the node classification task is to predict the disease into eight categories with class labels from \citep{hne}. The target of link prediction is to predict the existence of edges between genes and diseases. 
\end{itemize}

\begin{table}
\centering
\caption{Data Split of Node Classification Datasets.}\label{tab:benchmark split}
\small
\setlength{\tabcolsep}{0.9mm}
\begin{tabular}{ccccc}
\toprule
  & \#Nodes & \#Training  &\#Validation &\#Testing\\ 
 \midrule
DBLP          & 26,128  & 974  & 243  &   2,840            \\
IMDB          & 21,420  &  1,097  & 274  &  3,159  \\
ACM           & 10,942  &  726 & 181  &  2,118   \\
Freebase      & 180,098   & 1,909  & 477  & 4,446   \\
PubMed\_NC      & 63,109   &  295 & 73  &  86  \\ 
\bottomrule
\end{tabular}
\end{table}

\begin{table*}[!t]
\caption{The searched hyper-parameters for \ours with best performances on various datasets.}
\label{tab:searched hypers}

\centering
\small
\resizebox{1\textwidth}{!}{
\begin{tabular}{c ccccc ccccc cc}
\hline
Datasets & AttDim& lr&  wd  &dropAttn   &   droptFeat& featType& hiddenDim  & Layers&Epoch   &Heads& Decoder   & EdgeDim \\ \hline
DBLP & 32 &  1e-4 &1e-3&0.5 &0.5& 1& 64& 4& 300&8 & - & 64 \\ \hline
IMDB   & 3  &   1e-4  &  1e-3& 0.2 & 0.8 & 1 & 128 & 3 & 300 & 8 & - & 64  \\ \hline
ACM & 32  &   1e-3  &  1e-4& 0.8 & 0.8 & 1 & 64 & 2 & 300 & 8 & - & 64 \\ \hline
pubmed\_NC & 3  &   5e-3  &  1e-3& 0.8 & 0.2 & 1 & 128 & 2 & 300 & 8 & - & 64  \\ \hline
Freebase & 8  &   5e-4  &  1e-3& 0.5 & 0.5 & 2 & 16 & 2 & 300 & 8 & - & 0  \\ \hline
LastFM  & 64  &   5e-4  &  1e-4& 0.9 & 0.2 & 2 & 64 & 8 & 1000 & 2 & DotProduct & 64  \\ \hline
pubmed\_LP  & 32  &   1e-3  &  1e-4& 0.5 & 0.5 & 2 & 64 & 4 & 1000 & 2 & DistMult & 64  \\ \hline
\end{tabular}
}
\end{table*}

\subsection{Node Classification Data Split} \label{app:datasplit}
Following \citep{lv2021simple}, we have the training and validiation ratio 8:2, while keeping fixed testing test, and the statistics of testing data, as well as training and  validation data are listed in Table \ref{tab:benchmark split}.

\subsection{Additional Visualization Results}\label{app:visual}

\begin{figure}
    \centering
    \includegraphics[width=.38\textwidth]{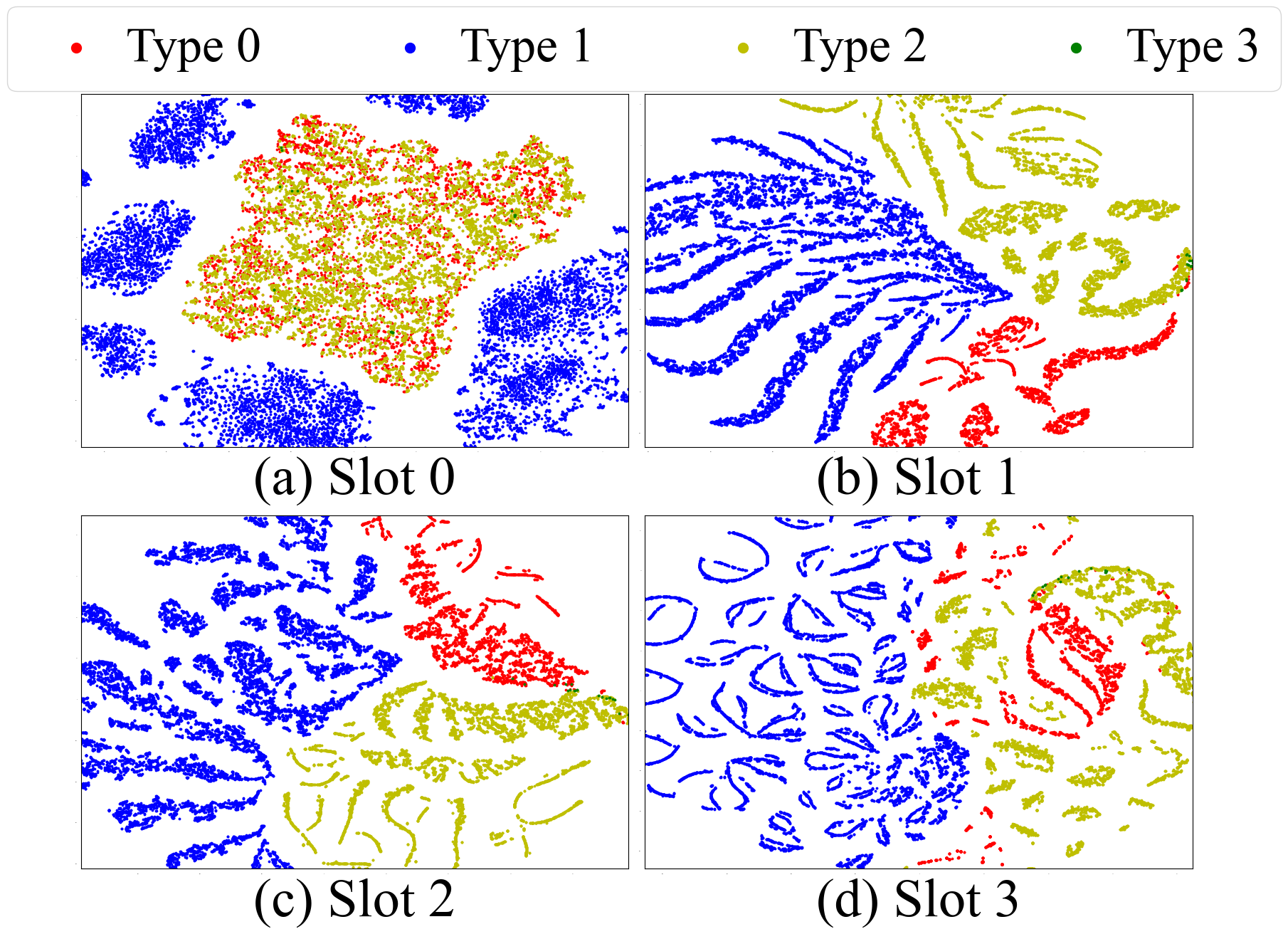}
    \caption{Tsne visualization on all slot representations of $3$-rd \ours layer on DBLP}\label{fig:tsneDBLP}
\end{figure}

Figure \ref{fig:tsneDBLP} reports the Tsne visualization on the slot representations of the 3-rd \ours layer on DBLP. 
In particular, Figure \ref{fig:tsneDBLP}(a) is the visualization of the 0-th slots 
(in type-0 feature space)
of all nodes with colors representing their own types, 
and Figure \ref{fig:tsneIMDB}(b), (c), (d) are the visualization of 1-st slots (in type-1 feature space), 
2-nd  slots (in type-2 feature space),
3-rd slots (in type-3 feature space)
of all nodes respectively. 
First, observe that the 4 slots of a node (for all nodes in a graph) learn very different representations, as the visualizations in the four figures exhibit different patterns. 
Note that the number of nodes in type 3 is very few compared to other three node types, and thus are nearly invisible in these figures. Therefore, we focus the discussion on nodes of the other three types in red, blue and yellow. 
In particular, as shown in Figure \ref{fig:tsneDBLP}(a), the 0-th slots (in type-0 feature space) are able to distinguish nodes in type 1 (blue) from others, but cannot distinguish nodes of types 0, 2, 3. 
On the other hand, in Figure \ref{fig:tsneDBLP}(b), the 1-st slots (in type-1 feature space) are able to distinguish nodes of different types in 0, 1, 2.
Similar observation can be made in Figures \ref{fig:tsneDBLP}(c),(d). 
The visualization in Figure \ref{fig:tsneDBLP} proves that \ours is able to learn richer semantics in finer granularity and also preserve the semantic differences of different node-type feature spaces into different slots.
This visualization also provide evidence on the superior performance of \ours.

\subsection{Link Prediction Decoders}\label{app:decoders}
Given $\bH_{\cv}$ and $\bH_{\cu}$, dot product decoder first computes dot product and then applies a sigmoid activation function: $Decoder(\bH_{\cv},\bH_{\cu})=\sigmoid(\innerPro{\bH_{\cv}}{\bH_{\cu}})$. 

DistMult requires a learnable square weight matrix $\bW_{\psi(\cv,\cu)}\in\real^{d_{L}\times d_{L}}$ for edge type $\psi(\cv,\cu)$.Then a bi-linear form between $\bH_v,\bH_u$ is calculated as: $Decoder(\bH_{\cv},\bH_{\cu})=\sigmoid(\bH_{\cv}^\top \bW_{\psi(\cv,\cu)}\bH_{\cu})$. 
The choice of dot product or DistMult is regarded as a part of hyper-tuning process \citep{lv2021simple}.

\subsection{Searched Hyper-parameters}\label{app:param}
To facilitate the re-producibility of this work, we present  the searched  hyper-parameters of \ours in Table~\ref{tab:searched hypers}  on different datasets. 
Note that due to the high computational cost in dataset Freebase, we set  the dimension $d_e=0$ of attention vector for edge types in this dataset, while $d_e=64$ in all other datasets. 
Moreover, three tricks are used in the implementation of \ours as suggested in~\citep{lv2021simple}: residual connection, attention residual and using hidden embedding in middle layers for link prediction tasks.

\subsection{Metrics}\label{sec:computation of metrics}\label{app:metrics}
Here we illustrate how we compute the four metrics: Macro-F1, Micro-F1, ROC-AUC and MRR. 

\textbf{Macro-F1}: The macro F1 score is computed using the average of all the per-class F1 scores.
Denote $\text{Precision}_{c}$ as the precision of class $c$, and $\text{Recall}_{c}$ as the recall of class $c$. We have
    \begin{equation}
        \begin{aligned}
            \text{Macro-F1}=\frac{1}{C}  \sum_{c\in [C]} \frac{2\text{Precision}_{c}\cdot\text{Recall}_{c}}{\text{Precision}_{c}+\text{Recall}_{c}}.
        \end{aligned}
    \end{equation}
    
    \noindent \textbf{Micro-F1}: The Micro-F1 score directly uses the total precision and recall scores. With the Precision and Recall scores of all nodes regardless of their classes, we have
    \begin{equation}
        \begin{aligned}
            \text{Micro-F1}=\frac{2\text{Precision}\cdot\text{Recall}}{\text{Precision}+\text{Recall}}.
        \end{aligned}
    \end{equation}
    
    \noindent \textbf{ROC-AUC}: The Area Under the Curve (AUC) score is calculated from Receiver Operating Characteristic (ROC).  Define the True Positive Rate (TPR) as  
 $        \text{TPR}=\frac{\text{TP}}{\text{TP+FN}}$,
    where TP, FN represent true positive and false negative respectively. The False Positive Rate (FPR) is calculated as $        \text{FPR}=\frac{\text{FP}}{\text{FP+TN}}$,
    where FP, TN represent false positive and true negative respectively.  One can see that, TPR and FPR scores vary if we change the classification threshold. 
    One can also prove that, there exists a one-to-one function $f$ such that under every threshold: $f(\text{FPR})=\text{TPR}$. 
    Then the ROC-AUC score is the area under the graph of this function $f$:
    \begin{equation}
        \text{ROC-AUC}=\int_{x\in[0,1]} f(x).
    \end{equation}

    \noindent \textbf{MRR}: Mean reciprocal rank (MRR) is a   metric usually used in recommendation system. Here we use it to evaluate performance in link prediction   as in~\citep{lv2021simple}. For each node $\cv$ involved in concerned edges, \ie positive and negative edges to be evaluated, one can sort its all related nodes $\cu$, \ie connected by these positive or negative nodes, by the similarities of their computed embeddings. Then, denote the minimal rank of the first occurred nodes $\cu$ connected with positive links $(\cv,\cu)\in\gE_{pos}$ in the above sorted nodes list as $R_{\cv}$, we have the MRR as the mean of reciprocal of this minimal rank over all the involved nodes $\{(\cv,\cu)\in\gE_{pos}\text{ or }\gE_{neg} \}$.

\subsection{Meta-paths Used in Baselines}
We provide the meta-paths used in baselines in  Table~\ref{tab:meta-paths}, which are widely adopted in ~\citep{lv2021simple,magnn,han}.

\begin{table}[!t]

\centering
\caption{Meta-paths used in our experiments.}
\label{tab:meta-paths}
\small
\begin{tabular}{ccc}
\hline
Dataset & Meta-paths & Meaning \\ \hline
DBLP & \makecell[c]{APA, APTPA, \\ APVPA} & \makecell[c]{A: Author, P: Paper,\\ T: Term, V: Venue} \\ \hline
IMDB & \makecell[c]{MDM, MAM,\\ DMD,  DMAMD,\\ AMA, AMDMA, \\ MKM} & \makecell[c]{M: Movie, D: Director, \\ A: Actor, K: Keyword} \\ \hline
ACM & \makecell[c]{PAP, PSP,\\ PcPAP,  PcPSP, \\PrPAP,  PrPSP,\\ PTP} & \makecell[c]{P: Paper, A: Author, \\ S: Subject, T: Term, \\ c: citation relation, \\ r: reference relation } \\ \hline
Freebase & \makecell[c]{BB, BFB, \\ BLMB, BPB, \\ BPSB, BOFB, \\ BUB} & \makecell[c]{B: Book, F: Film, \\ L: Location, M: Music, \\ P: Person, S: Sport, \\ O: Organization,\\ U: bUsiness} \\ \hline
LastFM & \makecell[c]{UU, UAU, \\ UATAU, AUA, \\ ATA, AUUA} & \makecell[c]{U: User, A: Artist, \\ T: Tag} \\ \hline
PubMed & \makecell[c]{DD, DGGD, \\ DCCD, DSSD} & \makecell[c]{D: Disease, G: Gene, \\ C: Chemical, S: Species} \\ \hline
\end{tabular}
\end{table}

\subsection{Proof of Theorem~\ref{thm:main}}\label{sec:proof of main}

\begin{proof}

The indication vector for the $i$-th CC  is denoted by $\mathbf{1}^{(i)}\in\mathbb{R}^n$. This vector indicates whether a node is in the maximal connected component $CC_i$, i.e.,
\begin{equation}
\small
    \mathbf{1}^{(i)}_j=\left\{
    \begin{array}{l}
        1, v_j \in CC_i \\
        0, v_j \not\in CC_i
       \end{array} \right.
\end{equation}
    If a graph has no bipartite components, the eigenvalues are all in [0,2) \cite{chung1997spectral}. The eigenspace  of $L_{rw}$ corresponding to eigenvalue 0 is spanned by $\{\mathbf{1}^{(i)}\}_{i=1}^{k}$ \cite{von2007tutorial}. 
    For $\alpha\in(0,1]$, the eigenvalues of $(I-\alpha L_{rw})$   fall into (-1,1], and the eigenspace of eigenvalue 1 is spanned by $\{\mathbf{1}^{(i)}\}_{i=1}^{k}$.

Denote $ (I-\alpha L_{rw}) =G=P \text{diag}\{\lambda_1,\lambda_2,...,\lambda_n\}P^\top $ as the diagnolization of symetric matrix $G$, where $\lambda_i$ is the eigenvalues and the $i$-th column of matrix $P$ is the corresponding eigenvector.
We compute 
\begin{equation}
\begin{aligned}
&[\lim_{\ell\to+\infty}{(I-\alpha L_{rw})}^\ell \mathbf{X}^{(t)}]_i\\
=&[\lim_{\ell\to+\infty}G^\ell \mathbf{X}^{(t)}]_i  \\
=&  [P \text{diag}\{\lim_{\ell\to+\infty}\lambda_1^\ell,\lim_{\ell\to+\infty}\lambda_2^\ell,...,\lim_{\ell\to+\infty}\lambda_n^\ell\}P^\top \mathbf{X}^{(t)}]_i
\end{aligned}
\end{equation}
Since all eigenvalues $\lambda_i\in(-1,1]$, only the eigenvalues equal to one could avoid shrinking to $0$. Thus,
\begin{equation}
    \begin{aligned}
        &[P \text{diag}\{\lim_{\ell\to+\infty}\lambda_1^\ell,\lim_{\ell\to+\infty}\lambda_2^\ell,...,\lim_{\ell\to+\infty}\lambda_n^\ell\}P^\top \mathbf{X}^{(t)}]_i\\
        =&  \sum_{j=1}^{n}\sum_{o=1}^{n}(\lim_{\ell\to+\infty}\lambda_o^\ell)P_{io}P_{jo}\mathbf{x}^{(t)}_j\\
        =&  \sum_{j=1}^{n}\sum_{1\leq o\leq n,\atop \lambda_o=1}P_{io}P_{jo}\mathbf{x}^{(t)}_j\\
    \end{aligned}
\end{equation}
Since the eigenspace of matrix $G$ corresponding to eigenvalue $1$  are spanned by $\{\mathbf{1}^{(i)}\}_{i=1}^{k}$~\citep{von2007tutorial}, and all of the eigenvectors within the matrix $P$'s columns are orthogonal to each other, we can   see that each column $P_{:,o}$ corresponding to eigenvalue $1$ (the count of this kind of vectors is equal to $k$) will be one of the vectors in the set $\{\frac{1}{\sqrt{|CC_i|}}\cdot\mathbf{1}^{(i)}\}_{i=1}^{k}$. These $k$ vectors all have $2$-norm  equal to $1$, where $|CC_i|$ is the size of the $i$-th CC.  
Thus if $\lambda_o=1$, $P_{:,o}$ is one of the vectors in set $\{\frac{1}{\sqrt{|CC_i|}}\mathbf{1}^{(i)}\}_{i=1}^{k}$. 
Then $P_{io}P_{jo}=\frac{1}{|cc(i)|}\iff \cv_j\in cc(i)$ (node $\cv_i$ and node $\cv_j$ in the same CC), where $|cc(i)|$ is the size of the CC which node $i$ is in, otherwise $P_{io}P_{jo}=0$. Therefore,
\begin{equation}\label{eq:proofMain0}
\small
\begin{aligned}
    [\lim_{\ell\to+\infty}G^\ell \mathbf{X}^{(t)}]_i=&\sum_{j=1}^{n}\sum_{1\leq o\leq n,\atop \lambda_o=1}P_{io}P_{jo}\mathbf{x}^{(t)}_j\\
    =& \sum_{\cv_j\in cc(i)}\frac{1}{|cc(i)|} \mathbf{x}^{(t)}_j.
\end{aligned}
\end{equation}
Since only nodes with type $t$ have non-zero features, Eq.~\ref{eq:proofMain0} is re-written as:
\begin{equation}\label{eq:proofMain1}
\small
\begin{aligned}
    [\lim_{\ell\to+\infty}G^\ell \mathbf{X}^{(t)}]_i
    =& \sum_{\cv_j\in cc(i)}\frac{1}{|cc(i)|} \mathbf{x}^{(t)}_j\\
    =& \sum_{\cv_j\in cc(i),\atop \phi(\cv_j)=t}\frac{1}{|cc(i)|} \mathbf{x}^{(t)}_j   .
\end{aligned}
\end{equation}

\end{proof}


\end{document}